\documentclass[twoside,11pt]{article}

\usepackage{jmlr2e}

\usepackage{amsmath}
\usepackage{amssymb}
\usepackage{multirow}
\usepackage{xcolor}
\usepackage{makecell}
\graphicspath{{figures/}}
\usepackage{listings}
\usepackage{color}

\definecolor{dkgreen}{rgb}{0,0.6,0}
\definecolor{gray}{rgb}{0.5,0.5,0.5}
\definecolor{mauve}{rgb}{0.58,0,0.82}

\lstdefinestyle{Python}{
    language        =   Python,
    captionpos      =   b,
    keywordstyle    =   \color{blue},
    keywordstyle    =   [2] \color{teal},
    stringstyle     =   \color{magenta},
    commentstyle    =   \color{red}\ttfamily,
    breaklines      =   true,
    columns         =   fixed,
    basewidth       =   0.5em
}

\ShortHeadings{A Modularized Bilevel Optimization Library in Python for Meta Learning}{Yaohua Liu and Risheng Liu}
\firstpageno{1}

\begin{document}

\title{BOML: A Modularized Bilevel Optimization Library in Python for Meta Learning}

\author{\name Yaohua Liu \email liuyaohua@mail.dlut.edu.cn \\
	\name Risheng Liu \email rsliu@dlut.edu.cn \\
	\addr International School of Information Science and Engineering, Dalian University of Technology\\
Key Laboratory for Ubiquitous Network and Service Software of Liaoning Province \\
Dalian, Liaoning, P. R. China
}

\maketitle

\begin{abstract}
Meta-learning (a.k.a. learning to learn) has recently emerged as a promising paradigm for a variety of applications. There are now many meta-learning methods, each focusing on different modeling aspects of base and meta learners, but all can be (re)formulated as specific bilevel optimization problems. This work presents BOML, a modularized optimization library that unifies several meta-learning algorithms into a common bilevel optimization framework. It provides a hierarchical optimization pipeline together with a variety of iteration modules, which can be used to solve the mainstream categories of meta-learning methods, such as meta-feature-based and meta-initialization-based formulations. The library is written in Python and is available at \url{https://github.com/dut-media-lab/BOML}.
\end{abstract}


\section{Introduction}
Meta-learning is the branch of machine learning that deals with the problem of ``learning to learn'' and has recently emerged as a potential learning paradigm that can gain experience over previous tasks and generalize that experience to unseen tasks proficiently. The applications of meta-learning span from few-shot classification~\citep{Reptile2018, MTNet2018}, and deep reinforcement learning~\citep{MAML2017,iMAML2019}, to neural architecture search~\citep{DARTS2019, Survery2020}.
However, due to the complex learning paradigms and the problem-specific formulations (for example, use different strategies to construct the base and meta learners), it actually requires significant optimization expertise to design efficient algorithms to solve these meta-learning problems.

In this work, by formulating meta-learning tasks from the bilevel optimization perspective, we establish a unified and modularized library, named BOML, for different categories of meta-learning approaches. Specifically, in BOML, we support two main categories of meta-learning paradigms, including meta-initialization-based~\citep{MAML2017} and meta-feature-based~\citep{franceschi2018bilevel} and implement a variety of recently developed bilevel optimization techniques, such as  Reverse Hyper-Gradient (RHG)~\citep{Reverse2017}, Truncated RHG (TRHG)~\citep{Truncated2019}, Meta-SGD~\citep{MetaSGD2017}, MT-net~\citep{MTNet2018}, WarpGrad~\citep{WarpGrad2020}, HOAG~\citep{Implicit2016} and Bilevel Descent Aggregation (BDA)~\citep{BDA2020}, for solving the meta-learning problems. Several first-order approximation schemes, including First-Order MAML (FMAML)~\citep{MAML2017} and DARTS~\citep{DARTS2019}, are also integrated into our BOML.

The key features of BOML can be summarized as follows: It provides a unified bilevel optimization framework to address different categories of existing meta-learning paradigms, offers a modularized algorithmic structure to integrate a variety of optimization techniques, and is flexible and extensible for potential meta-learning applications. We implement continuous code integration with \emph{Travis~CI} and \emph{Codecov} to obtain high code converge (more than 98\%). We also follow \emph{PEP8} naming convention to guarantee the code consistency. The documentations are developed with \emph{sphinx} and rendered using \emph{Read the Docs} (available at \url{https://boml.readthedocs.io}).

\section{A Unified Bilevel Optimization Paradigm for Meta Learning}
We first present a general bilevel optimization paradigm to unify different types of meta-learning approaches. Specifically, we define the meta data set as
$\mathcal{D}=\{\mathcal{D}^i \}_{i=1}^N$, where
$\mathcal{D}^i=\mathcal{D}^i_{\mathtt{tr}}\cup\mathcal{D}^i_{\mathtt{val}}$ is linked to the $i$-th task and $\mathcal{D}^i_{\mathtt{tr}}$ and $\mathcal{D}^i_{\mathtt{val}}$ respectively denote the training and validation sets. We denote the parameters of the base-learner as $\mathbf{y}^i$ for the $i$-th task. Then the meta-learner can be thought of as a function that maps the data set to the parameters of base-learner for new tasks, that is, $\mathbf{y}^i=\Psi(\mathbf{x},\mathcal{D}^i)$, where $\mathbf{x}$ is the parameter of the meta-leaner and should be shared across tasks.
With the above notations, we can formulate the general purpose of meta-learning tasks as the following bilevel optimization model:
\begin{equation}
\min\limits_{\mathbf{x}} F(\mathbf{x},\{\mathbf{y}^i\}_{i=1}^N), \quad s.t.
\quad \mathbf{y}^i\in\arg\min\limits_{\mathbf{y}^i}f(\mathbf{x},\mathbf{y}^i), \ i=1,\cdots,N,\label{eq:bo}
\end{equation}
where $f(\mathbf{x},\mathbf{y}^i)=\ell(\mathbf{x},\mathbf{y}^i,\mathcal{D}^i_{\mathtt{tr}})$ and $F(\mathbf{x},\{\mathbf{y}^i\}_{i=1}^N)=1/N\sum_{i=1}^N\ell(\mathbf{x},\mathbf{y}^i,\mathcal{D}^i_{\mathtt{val}})$ are called the Lower-Level (LL) and Upper-Level (UL) objectives, respectively. Here $\ell$ denotes task-specific loss functions such as cross-entropy. By reformulating the optimization process of $\mathbf{y}^i$ (with fixed $\mathbf{x}$) as a dynamical system, that is, $\mathbf{y}_0^i=\Psi_0(\mathbf{x},\mathcal{D}^i)$ and $\mathbf{y}_t^i=\Psi_t(\mathbf{x},\mathbf{y}_{t-1}^i,\mathcal{D}^i)$,
the meta-learner can be established as $\Psi=\Psi_T\circ\Psi_{T-1}\circ\cdots\circ\Psi_0$.

Based on the above construction, we can formulate different categories of meta-learning methods within the bilevel model in Eq.~\eqref{eq:bo}. Specifically, for meta-feature-based methods, we partition the learning model into a cross-task feature mapping (parameterized by $\mathbf{x}$) and task specific sub-models (parameterized by $\{\mathbf{y}^i\}_{i=1}^N$). While in meta-initialization-based approaches, we actually only need to consider $\mathbf{y}^i$ as the model parameters for the $i$-th task and set $\mathbf{x}$ as the initialization of the meta-learner.

\section{Design and Features of BOML}
In this section, we elaborate on BOML's design and features in three subsections.
\subsection{Optimization Process}

\begin{figure}[tbp]
  \centering
  \includegraphics[width=1.02\textwidth]{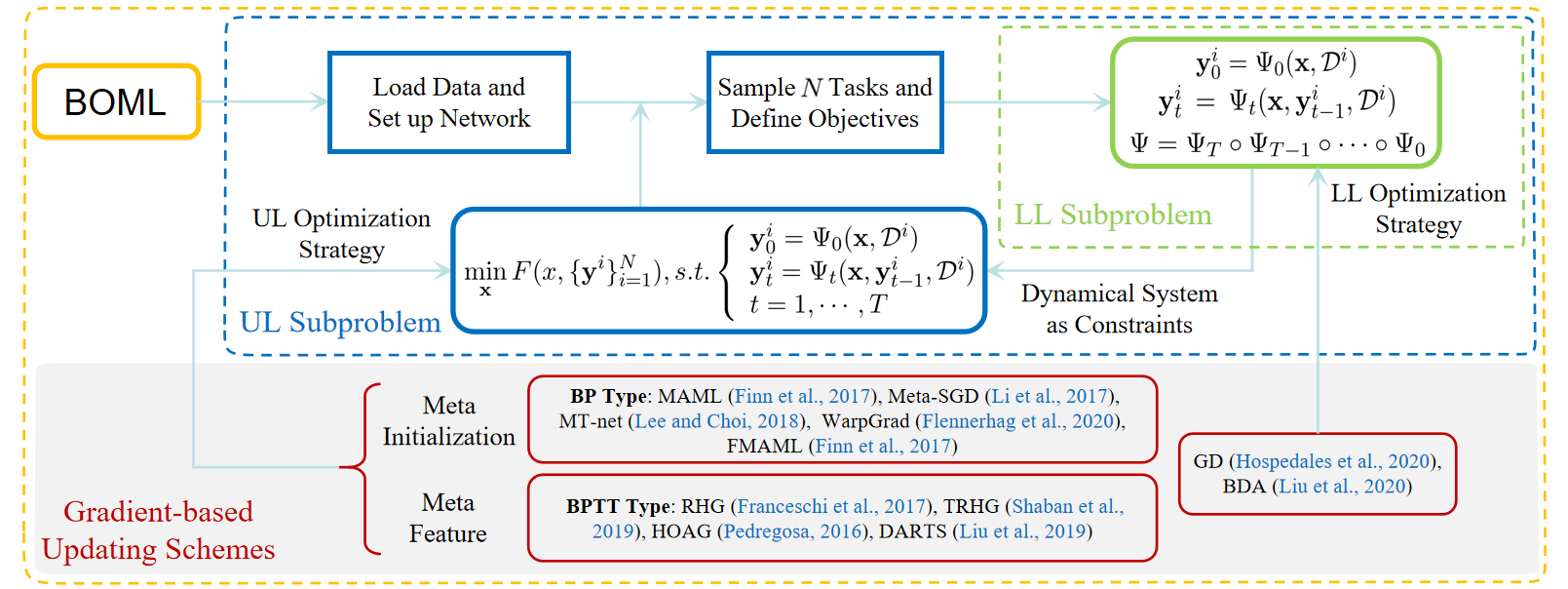}
  \caption{Illustrating the optimization process of BOML.}\label{optimization module}
\end{figure}

We first illustrate the optimization process of BOML for meta-learning in Figure~\ref{optimization module}. It can be seen that BOML constructs two nested optimization subproblems (blue and green dashed rectangles), which are respectively related to the LL variable $\mathbf{y}$ and UL variable $\mathbf{x}$ in Eq.~\eqref{eq:bo}. For the LL subproblem (w.r.t. $\mathbf{y}$), we can establish the dynamical system (parameterized by fixed $\mathbf{x}$) by performing Gradient Descent (GD) on the LL objective (that is, $f$). We also consider the recently proposed aggregation technique in BDA to integrate both the LL and UL objectives (that is, $f$ and $F$) to generate the dynamical system. As for the UL subproblem, we actually consider Back-Propagation (BP) and Back-Propagation-Through-Time (BPTT) \citep{baydin2017automatic} to calculate the gradients (with respect to $\mathbf{x}$) in meta-initialization-based and meta-feature-based tasks, respectively.\footnote{In fact, stochastic GD with momentum \citep{Survery2020} is used as our default updating rule and other mainstream schemes (for example, Adam~\citealp{kingma2014adam}) are also available in BOML.}

\subsection{Implementation Details}
BOML consists of the following six modules: $\mathtt{boml\_optimizer}$, $\mathtt{load\_data}$, $\mathtt{setup\_model}$, $\mathtt{lower\_iter}$, $\mathtt{upper\_iter}$ and $\mathtt{optimizer}$. BOML first instantiates $\mathtt{BOMLOptimizer}$ encapsulated in $\mathtt{boml\_optimizer}$ to store necessary parameters of model configuration and help manage the following procedures. (i) $\mathtt{load\_data}$ is invoked to process and sample batches of data for specific tasks. (ii) $\mathtt{setup\_model}$ defines network structure and initializes network parameters of meta-learner and base-learner on the basis of the data formats returned by $\mathtt{load\_data}$. (iii) $\mathtt{BOMLOptimizer}$ chooses built-in or extended strategies in $\mathtt{lower\_iter}$ and returns iterative formats of the dynamical system. (iv) $\mathtt{BOMLOptimizer}$ conducts the UL calculation with strategies in $\mathtt{upper\_iter}$, that calls $\mathtt{lower\_iter}$ during the back propagation process in turn. (v) To adapt to the nested gradient computation of dynamical systems in $\mathtt{lower\_iter}$ and $\mathtt{lower\_iter}$, BOML integrates different mainstream stochastic gradient updating schemes in $\mathtt{optimizer}$.

We consider few-shot classification~\citep{Fewshot2017} as our demo application and implement network structures for both meta-initialization-based and meta-feature-based approaches. Various loss functions (cross-entropy and mean-squared error) and regularization terms (L1 and L2 norms) are supported in BOML. We also provide loading configurations for several widely-used data sets, including  MNIST~\citep{Mnist1998}, Omniglot~\citep{Omniglot2011}, and MiniImageNet~\citep{Miniimagenet2017}. Listing~\ref{demo.py} presents a code snippet to demonstrate how to build bilevel optimization model for meta-learning. 

\begin{lstlisting}[
    style = Python,
    caption = {Code snippet of BOML on building bilevel optimization model for meta-learning.},
    label = {demo.py}
    ]
from boml import utils, load_data, extension
dataset = load_data.meta_omniglot(5, 1, 15) # load data 
ex = boml.BOMLExperiment(dataset) # standard class for experiment
bopt = boml.BOMLOptimizer("MetaInit", "Simple", "Simple") # create BOMLOptimizer
meta_learner = bopt.meta_learner(ex.x, dataset, "V1") # define meta-learner
model = boml_ho.base_learner(ex.x, meta_learner) # define base-learner
loss_inner = utils.cross_entropy(model.out, ex.y) # LL objective
inner_grad = bopt.ll_problem(loss_inner, learning_rate, T, var_list=model.var_list, experiment=ex) # LL subproblem
val_out = model.re_forward(ex.x_).out # output with validation set
loss_outer = utils.cross_entropy(val_out, ex.y_) # UL objective
bopt.ul_problem(loss_outer, meta_learning_rate, inner_grad, 
            extension.outer_parameters()) # UL subproblem
bopt.aggregate_all() # aggregate defined operations
\end{lstlisting}

\subsection{Comparison to Existing Libraries}

In the past few years, some libraries, such as Meta-Blocks\footnote{\url{https://github.com/alshedivat/meta-blocks.}} and Far-HO\footnote{\url{https://github.com/lucfra/FAR-HO.}}~\citep{Reverse2017}, have also been developed for meta-learning. Meta-Blocks is proposed in 2020 and still under development. This library stands out with its implementations for different modern meta-initialization-based algorithms. Far-HO is another library, which formulates meta-learning as specific hyper-parameter optimization task but is purely focused on meta-feature-based models. In contrast, BOML establishes a general bilevel optimization framework to unify both above two categories of meta-learning methods. We also implement a series of recently proposed extensions and accelerations in BOML. Table~\ref{tab:method} provides an extensive comparison of available algorithms for meta-learning
in these libraries.

\begin{table}[htbp]
    \centering
    \renewcommand\arraystretch{1.1}
	\setlength{\tabcolsep}{0.5mm}{
	\footnotesize
	\begin{tabular}{|c| c| c| c| c| c| c| c| c| c| c|}
		\hline
        Library & RHG & TRHG & HOAG & MAML & FMAML & MT-net  & Meta-SGD & WarpGrad &  DARTS & BDA \\
        \hline
        Meta-Blocks & - &- &- & $\checkmark$ & $\checkmark$ &- &- & - &- &- \\\hline
Far-HO& $\checkmark$ & $\checkmark$ & $\checkmark$ &- & - & -& - & - & - &- \\\hline
        BOML (Ours) & $\checkmark$ & $\checkmark$ & $\checkmark$ & $\checkmark$ & $\checkmark$ & $\checkmark$ &$\checkmark$ & $\checkmark$ & $\checkmark$ & $\checkmark$ \\
		\hline
	\end{tabular}}
	\caption{Availability of meta-learning algorithms published in the
		literature.}\label{tab:method}
\end{table}

\section{Conclusions and Future Works}

We (re)formulated mainstream meta-learning approaches into a unified bilevel optimization paradigm, and presented BOML to integrate common meta-learning approaches under our proposed taxonomy. For future work direction, we plan to continuously update the library to extend more gradient computation modules and related methods to support reinforcement learning strategies and neural architecture search.

\acks{Acknowledgments}
We would like to thank support from National Natural Science Foundation of China (Nos. 61922019 and 61672125), LiaoNing Revitalization Talents Program (XLYC1807088) and the Fundamental Research Funds for the Central Universities.
\vskip 0.2in

\bibliography{bibliography}

\end{document}